# Med-gte-hybrid: A contextual embedding transformer model for extracting actionable information from clinical texts

Aditya Kumar[1*†], Simon Rauch[2†], Mario Cypko[2], Oliver Amft[1,2]

[1] Hahn-Schickard, Georges-Köhler-Allee 302, Freiburg, 79110, Germany.

[2] Intelligent Embedded Systems Lab, University of Freiburg, Georges-Köhler-Allee 302, Freiburg, 79110, Germany.

*Corresponding author(s).

†These authors contributed equally to this work.

**Abstract**

We introduce a novel contextual embedding model *med-gte-hybrid* that was derived from the gte-large sentence transformer to extract information from unstructured clinical narratives. Our model tuning strategy for *med-gte-hybrid* combines contrastive learning and a denoising autoencoder. To evaluate the performance of *med-gte-hybrid*, we investigate several clinical prediction tasks in large patient cohorts extracted from the MIMIC-IV dataset, including Chronic Kidney Disease (CKD) patient prognosis, estimated glomerular filtration rate (eGFR) prediction, and patient mortality prediction. Furthermore, we demonstrate that the *med-gte-hybrid* model improves patient stratification, clustering, and text retrieval, thus outperforms current state-of-the-art models on the Massive Text Embedding Benchmark (MTEB). While some of our evaluations focus on CKD, our hybrid tuning of sentence transformers could be transferred to other medical domains and has the potential to improve clinical decision-making and personalised treatment pathways in various healthcare applications.

# 1 Introduction

Besides structured data, patient care information in Electronic Health Records (EHRs) comprises data in unstructured form (i.e., clinical notes). While EHR information may overlap between structured and unstructured sections, some crucial information remains only in the unstructured sections [1–5]. Relevant information for clinical decisions can easily be overlooked when dealing with large amounts of notes. Previous investigations have already found that clinical text alone can often provide sufficient information for decisions [2, 6, 7]. However, extracting actionable information from clinical text remains difficult due to language variability, inconsistent use of medical terminology, and lack of standardised formatting [8]. In addition, the lack of structure introduces ambiguities and inconsistencies in the text. Thus, it is still difficult to accurately interpret and analyse clinical notes for decision support systems. As a result, the development of advanced natural language processing (NLP) models that can extract and represent information from clinical narrative has become a focal point of research in digital medicine [2, 9, 10].

Contextual embedding models have emerged as a powerful tool for transforming unstructured texts into dense vector representations to encode rich semantic information [11, 12]. However, existing models often struggle with lengthy and complex medical documents. The issue is particularly pressing given the increasing volume of clinical data and the need for accurate information retrieval [13]. Current domain-specific embedding models, including ClinicalBERT [14] and BioBERT [15], have demonstrated effectiveness in specific clinical applications, but tend to lag behind generalist models in handling tasks, including semantic text similarity analysis [16, 17]. Notably, generalist models often outperform their clinical counterparts because of their broader exposure to various language contexts, resulting in enhanced understanding and retrieval capabilities [16–18]. However, a limitation of generalist models is their inability to effectively understand and process specialised medical terminology [17]. In particular, the precise meaning of medical text can vary profoundly depending on its context, which can lead to semantic misrepresentations. Consequently, the absence of a domain-specific model that is explicitly designed to process long-context tasks is an open challenge in the clinical domain.

Our work introduces a clinically specialised sentence transformer, termed *med-gte-hybrid*, designed for long-context tasks within the clinical domain (see Fig. 1). The model is based on gte-large [1] sentence transformer, an extension of the gte model [19]. For the gte-large model, self-supervised fine-tuning strategies have already been investigated. Here we combine for the first time, Simple Contrastive Learning (SimCSE) [20] and Transfomer-based Sequential Denoising Auto-Encoder (TSDAE) [21] to fine-tune the gte-large generalist model and derive the hybrid model *med-gte-hybrid*.With *med-gte-hybrid*, we aim to overcome the limitations of existing contextual embedding models, especially for long-context tasks in the clinical domain. This work makes the following contributions:

1. We propose a novel, specialised sentence transformer *med-gte-hybrid* that combines contrastive learning (SimCSE) with a denoising autoencoder (TSDAE) to interpret clinical notes of EHR and support clinical decisions. We derive two separate intermediate models *med-gte-simcse* and *med-gte-tsdae* and subsequently concatenate their embeddings to derive an ensemble that includes the complementary strengths of each intermediate model.

[1] https://huggingface.co/Alibaba-NLP/gte-large-en-v1.5

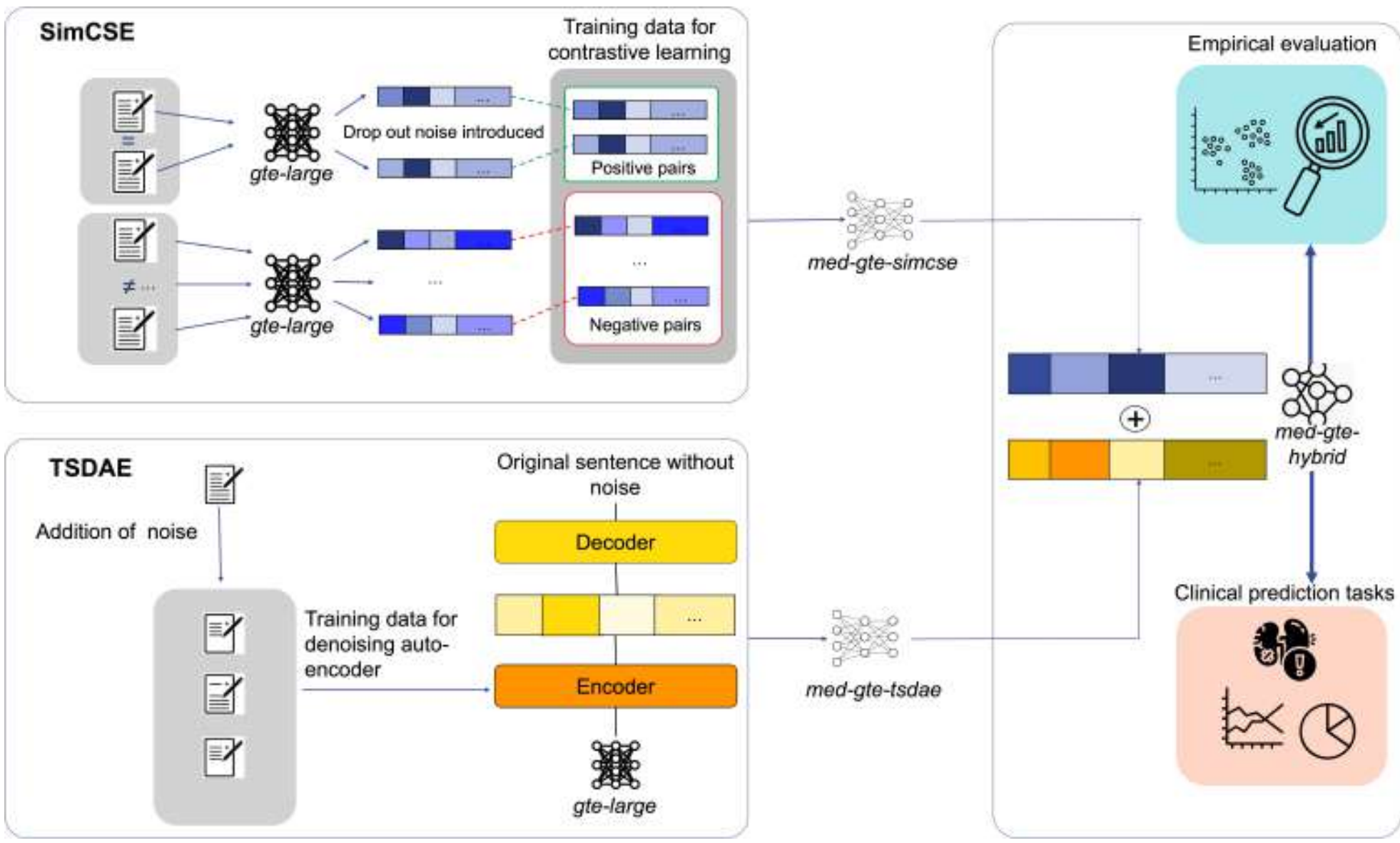

**Fig. 1**: Overview of the *med-gte-hybrid* model. The hybrid model combines two model training approaches that were adapted to the medical domain in this work: (1) Using Simple Contrastive Learning (SimCSE) led to the *med-gte-simcse* intermediate model, and (2) a Transfomer-based Sequential Denoising Auto-Encoder (TSDAE) led to the *med-gte-tsdae* intermediate model. *Med-gte-hybrid* concatenates embeddings of both individually trained intermediate models to create rich and robust representations of clinical text.

2. We demonstrate the capabilities of the *med-gte-hybrid* model in clinical prediction tasks that require narrative information, including Chronic Kidney Disease (CKD) patient prognosis, estimated glomerular filtration rate (eGFR) prediction, and patient mortality prediction. We compare the intermediate models and the hybrid model to gte-large and ClinicalBERT.
3. We show that *med-gte-hybrid* can be applied in patient stratification and clustering tasks to capture and represent clinical characteristics. In particular, we examine the new models on the Massive Text Embedding Benchmark (MTEB) and show that *med-gte-hybrid* outperforms state-of-the-art models.

## 2 Related works

Embedding models can be broadly classified into generalist models and domain-specific models. Popular generalist models include Bidirectional Encoder Representations from Transformers (BERT) [12], numerous variations of BERT [22–24], Jina embeddings [25], and Big Bird [26]. Generalist models are trained on diverse, large-scale datasets and are capable of handling a wide range of tasks across various domains. In contrast, domain-specific models, including clinical embedding models, are fine-tuned or trained on specialised datasets, enabling them to capture the semantic and contextual intricacies of domain-specific language, for example medical terminology in healthcare. Popular variations of BERT and BigBird are ClinicalBERT [14], BioBERT [15], PubMedBERT [27], Med-BERT [28], Clinical-Longformer [29], and Clinical-BigBird [30]. Med-BERT is designed for structured EHR. One major limitation of existing clinical models is their insufficient ability to process and understand the nuances of longer clinical texts (i.e. exceeding 2000 tokens). Many current models were developed primarily with shorter contexts in mind, and therefore struggle to maintain context over extended passages. Understanding patient reports is critical in healthcare settings, where detailed patient information is essential [31, 32].

In addition to the scope of training data used, embedding models can be further distinguished into word-embedding models and sentence transformers. Word embeddings capture the meaning of individual tokens by encoding them as dense vectors in a continuous space [10, 33]. In contrast, sentence transformers can handle multiple sentences at once (usually exceeding 2000 tokens), thus capture broader context. As a result, sentence transformers understand the complex relationships between words than word-embedding models. Sentence transformers facilitate more sophisticated downstream tasks, including sentence-level classification and semantic similarity [34]. Some models, including the Clinical-Longformer [29], can handle contexts up to 4096 tokens, but are not sentence transformers. Clinical-Longformer is a fine-tuned Longformer model, corresponding to a token-level encoder that produces one vector per token, albeit with an extended context window. Conversely, sentence transformers often employ a Siamese or dual-encoder setup that generates a single fixed-dimensional vector for an entire sentence or paragraph [34].

Current clinical embedding models lack contextual text embedding that is specifically designed for context tasks related to clinical notes. Excoffier et al. [17] compared word-embedding models in clinical short-context tasks, including sematic search, and found that generalist models outperform clinical models. While current clinical models rely primarily on token-level representations, Excoffier et al. further discussed the unexplored potential of clinically specialised sentence transformers that are explicitly designed to produce rich sentence-level embeddings.

Sentence transformers (e.g., gte-large-en-v1.5 with up to 8192 tokens) are designed to produce semantically meaningful embeddings at the sentence level [34]. With self-supervised learning, sentence transformers can adapt easily to clinical text, generalise across multiple clinical tasks [35, 36], and acquire meaningful representations without constraints of task-specific training [37].

Contrastive learning [38] and denoising autoencoder [39] are two notable self-supervised learning techniques that have been used to capture contextually rich and meaningful embeddings. Contrastive learning approaches have been used in applications, including multi-view medical report generation [40] and medical image analysis of knee X-ray images [41]. Denoising autoencoder approaches approaches have been used to achieve superior results in specific tasks, including semantic text similarity [42].

# 3 Methods

## 3.1 Base model selection

We attempt to balance computational efficiency, generalization, and performance, by focusing on models with 250-500 million parameters and an embedding dimension of 1024. Models in the above size range have sufficient complexity to capture relevant clinical features, while avoiding the computational overhead associated with larger models [43]. We focused on top-ranked state-of-the-art generalist models of the Massive Text Embedding Benchmark (MTEB) leaderboard on Hugging Face[2]. MTEB [44] provides a standardised comparison for embedding models across various tasks. We examined the top four public models as of June 2024: bge-large, uae-large-v1, mxbai-large-v1, and gte-large, all of which had demonstrated to scale across a range of tasks, including classification, clustering, retrieval, and semantic textual similarity. The gte-large model performed best and hence was selected as base model for fine-tuning on clinical texts.

## 3.2 Model fine-tuning

Traditional supervised learning methods are often not practical for fine-tuning due to a lack of labels and the amount of data needed. While supervised learning is used for specific classification tasks, the resulting fine-tuned model may not generalise across the text domain [45]. Domain adaptation using supervised fine-tuning is typically done for datasets with well-defined semantic structures [46]. In the present work, we employed two self-supervised approaches for fine-tuning, namely SimCSE and TSDAE.

***SimCSE***

SimCSE is an effective framework for self-supervised contrastive learning from unlabelled text corpora. The key idea is to create positive and negative pairs from the input sentences and use contrastive learning to train the model. Given a set of sentences $S = \{s_1, s_2, \ldots, s_n\}$, for each sentence $s_i$, positive pairs are created by applying dropout during the encoding process twice, generating two different embeddings for the same sentence, denoted as $z_i^{(1)}$ and $z_i^{(2)}$, which form a positive pair. The intuition is that despite the slight variation due to dropout, the model should learn that the two embeddings represent the same semantic meaning. For negative pairs, other sentences $s_j$ from the dataset $(i \neq j)$ are randomly sampled, and their embeddings $z_j$ are paired with $z_i^{(1)}$. Here, the intuition is that random sampling yields sentences of different meaning, thus representing meaningful examples of negative pairs. The model then uses InfoNCE loss [47], to maximise the similarity of positive pairs $\text{sim}(z_i^{(1)}, z_i^{(2)})$ and minimise similarity of negative pairs $\text{sim}(z_i^{(1)}, z_j)$, where $j \neq i$. The loss for a single sentence $s_i$ is expressed as:

$$L_i = -\log \frac{\exp(\text{sim}(z_i^{(1)}, z_i^{(2)})/\tau)}{\sum_j \exp(\text{sim}(z_i^{(1)}, z_j)/\tau)}$$

where $\tau$ is a parameter controlling the sharpness of the distribution. The model is trained by minimising the total contrastive loss across all sentences $L = \sum_i L_i$, where it learns to pull the embeddings of positive pairs closer together while pushing the embeddings of negative pairs further apart. The process enhances the quality of sentence representations, thus allowing the model to better capture semantic similarities and differences. In summary, SimCSE adapts pre-trained models to the specific characteristics of the text in the training data and enables them to generate versatile embeddings for different tasks.

[2] https://huggingface.co/spaces/mteb/leaderboard

***TSDAE***

TSDAE is a sentence transformer training method for domain adaptation. The approach is to treat the embedding model like a denoising auto-encoder, training it to reconstruct corrupted sentences. The process starts with a set of sentences $S = \{s_1, s_2, \ldots, s_n\}$. For each sentence $s_i$, a corruption function is applied, typically removing or masking a certain percentage of the words, resulting in a corrupted sentence $s_i^{corrupt}$. The model's task is to reconstruct the original sentence $s_i$ from this corrupted version. The encoder processes the corrupted sentence $s_i^{corrupt}$ and produces a dense embedding vector $z_i$, which captures the contextual information from the corrupted input. The embedding is then passed through a decoder that attempts to reconstruct the original sentence $s_i$ by predicting the missing or masked tokens. The reconstruction loss $L$ is minimised during training, typically using a cross-entropy loss function, as follows:

$$L = \sum_i \text{CrossEntropy}(s_i, \hat{s_i})$$

where $s_i$ is the original sentence, and $\hat{s_i}$ is the reconstructed sentence. By minimising cross-entropy loss, the model learns to generate embeddings that capture both local and global dependencies in the input and lead to better generalisation across linguistic patterns and domains. As the model learns to reconstruct corrupted sentences during training, it could generate more robust and meaningful sentence representations, which are useful for handling noisy or incomplete data as they are often encountered in clinical notes.

Given the noise, inconsistent terminology, and typos present in clinical notes, TSDAE is well-suited to adapt a model that should deal with clinical notes and also guide the model to enhance the robustness of the feature representations. Here, we follow the optimal configuration recommended by Wang et el. in the TSDAE paper, where sentences are corrupted by deleting words at a ratio of 0.6.

***Hybrid model***

Both training approaches, SimCSE and TSDAE, have some key differences, leading to intermediate models that learn different features and have distinct strengths. The intermediate model trained according to the SimCSE approach (*med-gte-simcse*) could recognise fine-grained differences in meaning across sentences, making it sensitive to subtle semantic variations. In contrast, the intermediate model trained using the TSDAE approach (*med-gte-tsdae*) could produce embeddings that are robust and rich in contextual information.

In *med-gte-hybrid*, the embeddings generated by both intermediate models *med-gte-simcse* and *med-gte-tsdae* were concatenated to form a unified representation. Thus *med-gte-hybrid* effectively leverages an ensemble approach to combine complementary strengths of each intermediate model. By concatenating the embeddings, we capture a wider range of features, as both, *med-gte-simcse* and *med-gte-tsdae*, learnt different feature subsets. Additionally, the combination of embeddings serves as a form of regularization, thus mitigates potential overfitting of the intermediate models.

# 4 Evaluation and clinical prediction tasks

## 4.1 Dataset and pre-processing

We utilised the MIMIC-IV v2.2 dataset [48] for both fine-tuning and evaluation purposes. For model fine-tuning, the MIMIC-IV-Note subset [49], comprising over 2 million unstructured and unlabeled clinical notes, was utilised. Each note is associated with a unique patient and admission identifier to precisely link to an individual patient and a specific hospital visit. Clinical notes were recorded during standard patient care.

In a pre-processing step, we removed line breaks and special characters, including "==" and "___", which were used to mask personal information. To segment texts into sentences, we used the sentence tokeniser of the nltk library [3]. The sentence tokeniser identified abbreviations and other contextual clues to decide whether a period ends a sentence or not. Since clinical notes often do not follow rigid grammar rules, sentence fragments with less than five words were removed. The aforementioned pre-processing preserves the structure of clinical notes and allows the sentence transformer models to generate robust embeddings.

## 4.2 Evaluation data subset

We extracted two further data subsets from the MIMIC-IV database for evaluation: the CKD cohort and the mortality cohort. We labelled the patients, e.g., with eGFR values in the CKD subset. For CKD prognosis, a cohort of 3932 patients diagnosed with CKD, amounting to 10,000 admission cases was extracted. An example of the extracted data is shown in Fig. 2. For mortality prediction, a separate cohort of 10,000 patients was extracted.

Furthermore, we employed non-MIMIC datasets to further evaluate performance across a variety of tasks. In particular, scientific medical and biomedical datasets from MTEB were used, which are similar to clinical notes in their use of medical terminology, but have a more rigid grammatical structure. In particular, we used BIOSSES, MedrxivS2S, MedrxivP2P, PubHealthQA, and MedQA datasets. Additionally, the Mayo dataset [50], which is not part of MTEB, was used to provide an additional evaluation benchmark in the clinical domain.

[3]https://www.nltk.org

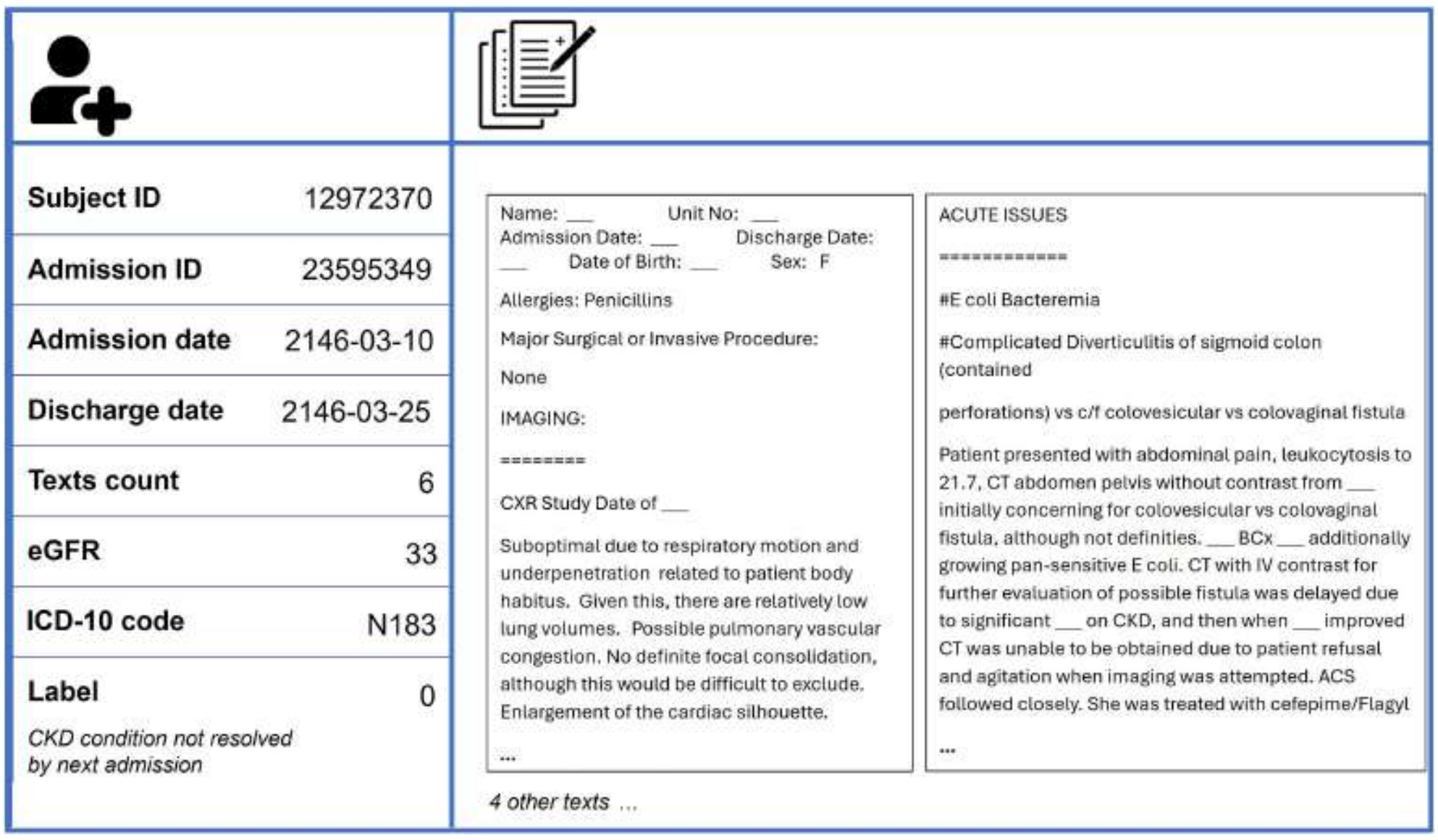

**Fig. 2**: Example of the clinical admission data, derived from one of the 3932 CKD patients of the MIMIC-IV database. A total of 10,000 admission cases were extracted for our analysis, including eGFR value, ICD-10 code, and prognosis label. For privacy reasons, MIMIC-IV admission and discharge dates were shifted into the future by a random offset.

### 4.3 Clinical prediction tasks

The capabilities of the *med-gte-hybrid* were explored across three distinct clinical prediction tasks: CKD prognosis, eGFR prediction, and mortality prediction. CKD prognosis and eGFR are closely related, as eGFR serves as a key prognostic indicator of CKD progression. The first two tasks demonstrate the dual capability of the *med-gte-hybrid* model in handling both, classification (CKD prognosis) and regression (eGFR prediction) tasks. In the eGFR prediction task, we analysed whether the model can infer eGFR values from clinical texts that do not explicitly mention eGFR, i.e., testing the ability of the model to capture contextual information. For the mortality prediction task, we included patients without CKD, to highlight the model's robustness and versatility in handling diverse healthcare prediction tasks. Furthermore, we investigated the capability of *med-gte-hybrid* model to cluster relevant features and support patient stratification by grouping similar patient profiles.

All clinical prediction tasks were evaluated using a 5-fold cross-validation with a stratified 80/20 train-test data split to maintain consistent class imbalance across all folds. Besides average performance, we report the standard deviation across folds to interpret model stability. Furthermore, we compared performance with ClinicalBERT and four state-of-the-art generalist models: bge-large, uae-large-v1, mxbai-large-v1, and gte-large.

### *CKD prognosis task*

We identified patients with CKD by filtering the MIMIC-IV database for admissions with ICD-10 [51] codes N181-N189 (see Tab. 1 for the ICD-10 code details). For each admission, we assigned a binary label: 0, if the CKD condition was not cured or resolved by the next admission of that patient, and 1, if the CKD condition was resolved.

| ICD-10 code | Description | Case count |
|---|---|---|
| **N181** | CKD, Stage 1 | 38 |
| **N182** | CKD, Stage 2 | 329 |
| **N183** | CKD, Stage 3 | 2824 |
| **N184** | CKD, Stage 4 | 846 |
| **N185** | CKD, Stage 5 | 213 |
| **N186** | End stage renal disease | 2120 |
| **N189** | Unspecified CKD | 3630 |

**Table 1**: ICD-10 codes for CKD stages that indicate chronic kidney disease severity. CKD patient count per stage considered in our analysis is shown.

We used the MIMIC-IV preprocessing pipeline developed by Gupta et al. [52] to create the CKD cohort and assign the corresponding labels. For the CKD prognosis task, all available notes for each admission were extracted and encoded by the proposed *med-gte-hybrid* model to generate one embedding per note (i.e. note-level embeddings). The note-level embeddings were then averaged into a feature vector for each admission. A feed-forward neural network with one hidden layer was trained to predict 0 or 1 based on the feature vector.

### *eGFR prediction*

The eGFR value is not explicitly mentioned in the clinical notes, which requires the regression model to infer the value from other embedded linguistic features. For patients in the CKD cohort, we extracted eGFR values from their EHR to serve as ground truth. Similar to the CKD prognosis, a feed-forward neural network was trained to output a continuous eGFR value from a feature vector provided by *med-gte-hybrid*.

### *Mortality prediction*

The mortality prediction analysis cohort was randomly selected 10,000 patients in the MIMIC-IV dataset. The resulting dataset was more diverse than the CKD prognosis dataset and therefore can offer complementary insight into the *med-gte-hybrid* performance. Patients were labelled as died (labelled as 1), if they died within 30 days after hospital discharge (1095 patients in the selected cohort). Mortality prediction was performed in the same way as the CKD prognosis.

### 4.4 MTEB evaluation

Two sentence similarity tasks were selected, which used Mayo and BIOSSES datasets to evaluate the models' ability to measure semantic textual similarity between clinical sentence pairs. The performance for the aforementioned tasks are reported using Spearman's correlation between annotated and model-derived cosine similarities. Spearman's correlation coefficient $\rho$ measures the rank correlation between annotated semantic similarities $y_i$ and cosine similarities of the model's embeddings $\hat{y}_i$:

$$\rho = 1 - \frac{6\sum_i d_i^2}{n(n^2-1)}$$

where $d_i$ is the difference between ranks of $y_i$ and $\hat{y}_i$, and $n$ is the number of sentence pairs.

Additionally, we evaluated two clustering tasks (using Medrxiv S2S and Medrxiv P2P datasets) for clustering quality using the V-measure [53]. The V-measure is the harmonic mean of completeness $C$ and homogeneity $H$, given by:

$$V = 2 \cdot \frac{C \cdot H}{C + H}$$

where completeness measures the extent to which all instances of the same class are assigned to the same cluster, and homogeneity measures how pure the clusters are (i.e. how many instances in each cluster belong to the same class).

Finally, we analysed the models' ability to retrieve relevant paragraphs in two retrieval tasks that used PublicHealthQA and MedicalQARetrieval datasets. We used the normalised discounted cumulative gain (nDCG@10) [54] to rank accuracy. nDCG@10 calculates the ranking quality of the first ten retrieved documents. nDCG is given by:

$$\text{nDCG@10} = \frac{1}{Z}\sum_{i=1}^{10}\frac{2^{\text{rel}_i}-1}{\log_2(i+1)}$$

where $\text{rel}_i$ is the relevance of the $i$-th document, and $Z$ is a normalisation factor to keep the score between 0 and 1. For top-ranked documents, nDCG values increase with relevance scores.

### 4.5 Model inspection

To better understand the nature of the embeddings produced by our fine-tuned models, we examined the embedding space. We applied UMAP to reduce the high-dimensional embedding vectors of each admission in the CKD cohort into a two-dimensional space for visualisation. Our aim was to investigate whether the embedding space clusters relevant clinical features and exhibits discernible patterns. Additionally, we employed SHapley Additive exPlanations (SHAP) to analyse classifier predictions. With SHAP we could identify features (i.e. tokens and phrases) that influenced the classifier's output.

# 5 Results

## 5.1 Clinical prediction tasks

***CKD prognosis task***

Results of the CKD prognosis task are shown in Fig. 3 (a,b). ClinicalBERT showed the lowest performance and was outperformed by the general embedding model gte-large with an Area Under the Receiver Operating Characteristic (AUROC) of 0.70 and an Area Under the Precision Recall Curve (AUPRC) of 0.52. *Med-gte-tsdae* yielded a larger AUROC of 0.75 and AUPRC of 0.58. The SimCSE fine-tuned model further enhanced performance to 0.81 and 0.67. Finally, *med-gte-hybrid* yielded the highest performance with 0.84 and 0.74 for AUROC and AUPRC respectively.

The general embeddings of gte-large capture more relevant features for CKD prognosis than ClinicalBERT, which was trained on clinical notes from MIMIC-III using standard BERT training tasks (Masked Language Modeling and Next Sentence Prediction). All of our fine-tuned models further refined the embeddings. In contrast to gte-large, the *med-gte-hybrid* model yielded 20.4% and 44.4% improvements in AUROC and AUPRC, respectively.

***eGFR prediction***

The results the eGFR prediction task are shown in Fig. 3 (c,d). Here, gte-large achieved a MAE of 11.6. *Med-gte-tsdae* reduces the error further to 11.4, while *med-gte-simcse* demonstrates a substantial improvement, with an MAE of 11.0. The *med-gte-hybrid* model again provides the best performance, with the lowest MAE of 10.5.

As a benchmark, an average eGFR value predictor with the average value derived across the entire patient cohort was defined. The average predictor yielded a MAE of 18.4. In particular, the *med-gte-hybrid* model provided more accurate predictions, indicating that the feature representations encode relevant information for eGFR, including, e.g., demographics, description of symptoms, and information about medication.

***Mortality prediction***

Fig. 3 (e,f) shows the performance for mortality prediction. We observe the same trend as for the other tasks: ClinicalBERT performed worst with an AUROC of 0.77 and an AUPRC of 0.30 and was outperformed by the gte-large model. The *med-gte-tsdae* model enhanced performance over gte-large with an AUROC of 0.85 and an AUPRC of 0.43. *Med-gte-simcse* exhibited an increase over *med-gte-tsdae* with AUROC of 0.87 and AUPRC of 0.47. The *med-gte-hybrid* model achieved the best performance scores with AUROC of 0.88 and AUPRC of 0.49.

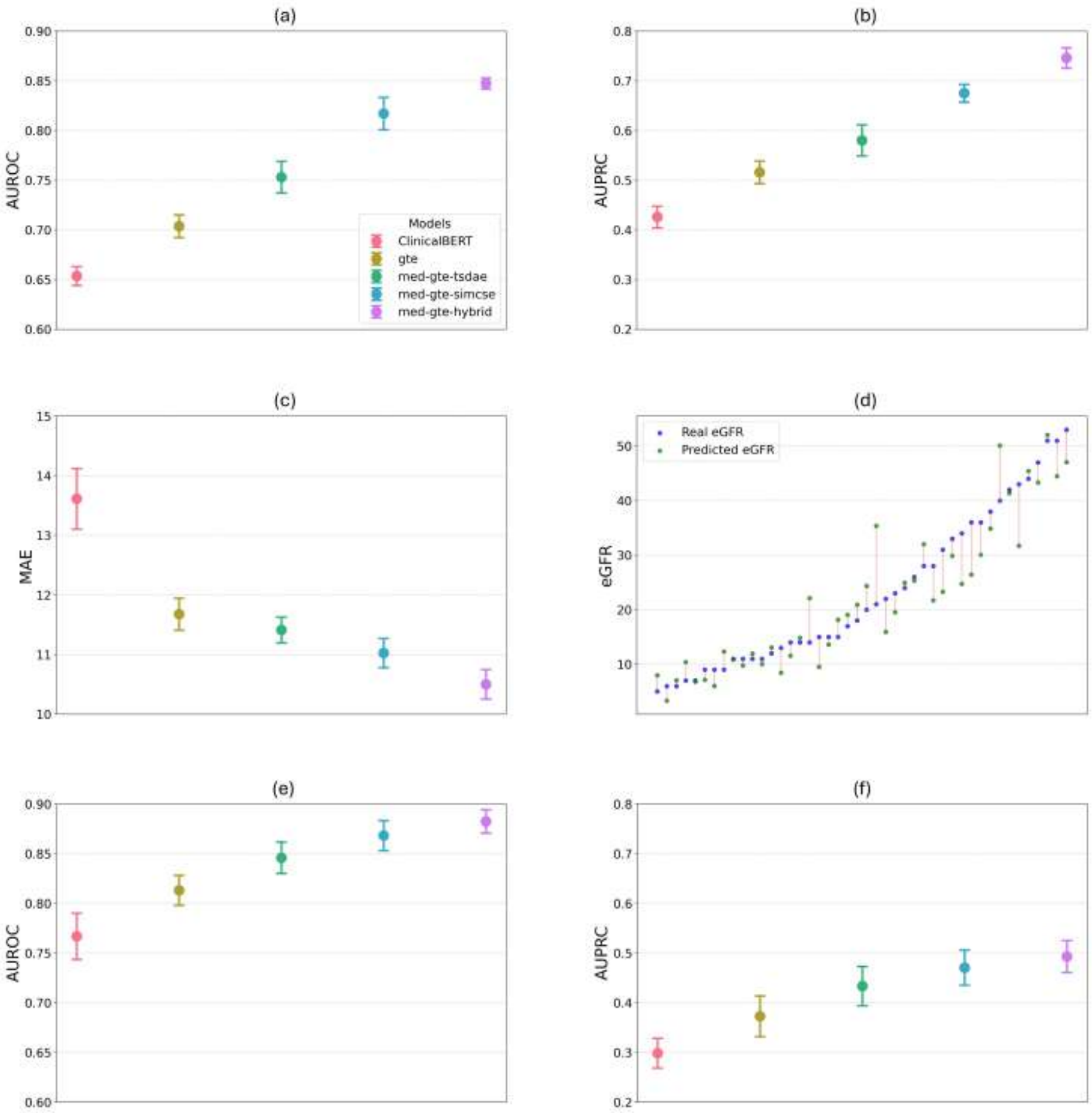

**Fig. 3**: Results of exemplary clinical prediction tasks (CKD prognosis, eGFR prediction and mortality prediction) for *med-gte-hybrid* and other benchmark models. (a) AUROC for CKD prognosis. (b) AUPRC for CKD prognosis. (c) MAE for eGFR prediction. (d) Sample of 50 eGFR predictions with *med-gte-hybrid* in comparison to real eGFRs, sorted by real eGFRs. (e) AUROC for mortality prediction. (f) AUPRC prediction for mortality prediction. *Med-gte-hybrid* yielded superior results for CKD prognosis with an AUROC of 0.85 and an AUPRC of 0.75 as well as in the eGFR prediction with a MAE of 10.50. In the mortality prediction task, *med-gte-hybrid* achieves the best performance with an AUROC of 0.88 and an AUPRC of 0.49.

The improvements seen with the fine-tuned models over the base model in mortality prediction are smaller compared to those observed in CKD prognosis. In particular, the AUPRC values for mortality prediction are lower compared to both the CKD prognosis task and the AUROC for mortality prediction.

### 5.2 MTEB evaluation

Tab. 2 presents the results of the MTEB evaluation. Scores of the four state-of-the-art embedding models (bge-large, uae-large, mxbai-large, and gte-large) were close to each other, with gte-large often scoring highest. Both of our single fine-tuned models outperformed state-of-the-art models across all tasks. *Med-gte-tsdae* performed the best in both retrieval tasks, while our *med-gte-hybrid* scores were highest in all other tasks.

| | Mayo | BIOSSES | MedrxivS2S | MedrxivP2P | PubHealthQA | MedQA |
|---|---|---|---|---|---|---|
| ClinicalBERT | 25.2 | 59.8 | 29.0 | 29.8 | 40.0 | 15.5 |
| **State-of-the-art embedding models** | | | | | | |
| bge-large | 65.1 | 88.4 | 31.4 | 32.5 | 77.6 | 70.1 |
| uae-large-v1 | 68.4 | 88.5 | 31.1 | 33.2 | 78.0 | 70.1 |
| mxbai-large-v1 | 69.7 | 88.2 | 31.6 | 33.4 | 79.1 | 71.1 |
| gte-large | 69.6 | 88.5 | 32.9 | 35.0 | 86.2 | 72.6 |
| **Fine-tuned models** | | | | | | |
| med-gte-tsdae | 71.4 | 89.4 | 35.1 | 37.2 | 88.0 | 73.8 |
| med-gte-simcse | 73.8 | 90.4 | 36.2 | 38.0 | 87.5 | 73.0 |
| med-gte-hybrid | 73.9 | 90.5 | 36.7 | 38.8 | 87.3 | 72.5 |

**Table 2**: Results of MTEB evaluation. Semantic textual similarity tasks Mayo and BIOSSES reported by Spearman r. Clustering tasks Medrxiv S2S/P2P reported by V-measure. Retrieval tasks PublicHealthQA and MedicalQARetrieval reported by nDCG@10. All metrics are reported in a 0-100 scale. The *med-gte-hybrid* model achieved the highest performance in Semantic textual similarity and clustering tasks, while *med-gte-tsdae* outperformed other models in retrieval tasks. Notably, the individual fine-tuned models already surpassed performance of the state-of-the-art embedding models.

### 5.3 Model inspection

The UMAP visualization for *med-gte-hybrid* are shown in Fig. 4 (c). The illustration revealed distinct clusters associated with specific ICD-10 codes within the CKD cohort. Notably, ICD-10 code N186 (end stage renal disease) formed a well-separated cluster, signalling a clear differentiation in the embedding space. In contrast, ICD-10 codes N183 (CKD stage 3) and N184 (CKD stage 4) exhibited visible, but less distinct clusters. Other ICD-10 codes showed more dispersed patterns, likely due to smaller sample sizes. Our findings indicate that the *med-gte-hybrid* model captures underlying clinical narratives within the embeddings and that there is a clear separation between specific ICD-10 codes in the embedding space, which indicates the model's proficiency in tasks related to patient stratification.

The SHAP analysis of a low-confidence prediction (p=0.49) revealed specific clinical features that played a role in the decision-making process of the model as shown in Fig. 4 (a, b). Key phrases, including "obstructing renal stone" and "multiple hypoattenuating lesions" were linked to a negative prognosis, while terms, including "stable" and "no appreciable pleural effusion" were associated with a more favourable outcome. The attributions align with clinical expectations and indicate that the embeddings capture meaningful clinical features for clinical prediction tasks.

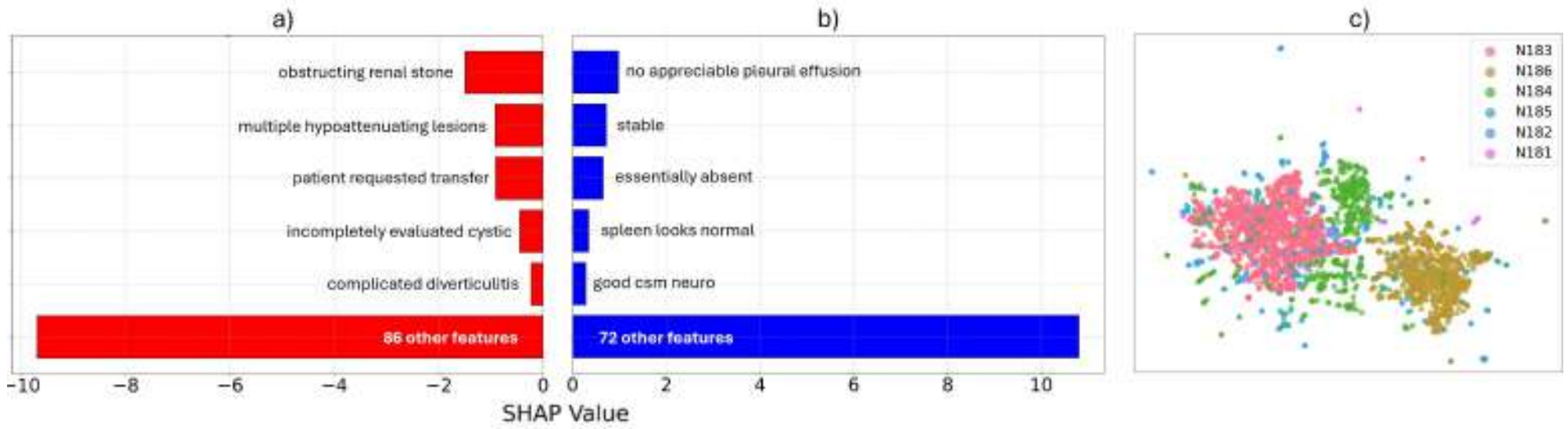

**Fig. 4**: **Empirical evaluation of *med-gte-hybrid* embeddings.** (a, b) SHAP analysis for a CKD prognosis sample: (a) Top five phrases increasing the probability of predicting 0, thus indicating negative influence. (b) Top five phrases increasing the probability of predicting 1, thus indicating positive influence. For instance, "obstructing renal stone" suggests unresolved CKD, while "no appreciable pleural effusion" indicates positive progression. (c) UMAP-reduced embedding vectors for CKD cohort admissions show distinct clusters that align with clinical expectations. SHAP analysis and clustering verify the meaningfulness of *med-gte-hybrid* embeddings.

## 6 Discussion

We investigated the *med-gte-hybrid* model and two intermediate models, *med-gte-simcse* and *med-gte-tsdae*, and compared them against state-of-the-art models and ClinicalBERT. Although prior work on Clinical-Longformer [29] reported superior performance relative to ClinicalBERT, we excluded it from our evaluation, because the released model lacks crucial pooling and layer selection details. Without clear instructions on deriving embeddings at sentence or document level, meaningful comparisons to our models were not feasible. The superior performance of *med-gte-hybrid* across our analyses indicates that our approach has the potential to leverage clinical texts to improve patient stratification and predictive outcomes. In particular, *med-gte-hybrid* showed excellent performance for CKD care. In CKD, clinical narratives frequently include a substantial amount of unstructured data, including prescriptions, physician notes, medical histories, discharge summaries, and treatment plans [55].

*Med-gte-hybrid* effectively captured the underlying clinical narratives, in particular for well-defined clinical conditions, including end-stage renal disease. CKD Stages 3 and 4 showed less clear clustering, which can be attributed to more varied and subtle clinical characteristics. Nevertheless, the clear separation of conditions in the embedding space demonstrates the proficiency of *med-gte-hybrid* in tasks that require patient stratification. Additionally, SHAP analysis of model predictions provided insights into the decision-making process, with key clinical features, including terms related to renal function and stability that align with expected outcomes.

For CKD prognosis, our results revealed that the generalist embedding model gte-large captured more relevant features than ClinicalBERT, where the latter was trained using traditional BERT techniques. Fine-tuning of gte-large further enhanced the feature representation. Our findings highlight the importance of adapting embedding models to specific clinical contexts, where context-specific nuances play a crucial role in model effectiveness. In particular, for CKD prognosis, the fine-tuned models better capture narrative information from detailed clinical notes, resulting in improved prediction of patient outcomes.

In the eGFR prediction task, *med-gte-hybrid*, once again, delivered the best performance, demonstrating the ability to encode clinically relevant features, including patient demographics, symptoms, and medication information. eGFR values were not present in the clinical notes, thus the model inferred key clinical metrics from text only. The strong predictive performance of *med-gte-hybrid* highlights the potential to enhance healthcare analytics, particularly in long-context narratives. The model's capability to capture hidden patterns between medical conditions, treatments, and patient outcomes, further supports its utility in clinical decision-making when explicit data may be limited or unavailable.

In the mortality prediction task, *med-gte-hybrid* demonstrated performance improvements, however, the gains were less profound compared to those achieved in CKD prognosis and eGFR prediction tasks. The mortality prediction task is highly heterogeneous and depends on a broad range of clinical features. Reduced AUPRC compared to AUROC may be related to class imbalance in the data [56, 57]. The dataset contained notably fewer positive cases (i.e., patients, who died within 30 days after discharge), which may affect the model's ability to accurately predict mortality. The distribution of positive to negative cases was 1,095 to 8,905.

In retrieval tasks, the intermediate *med-gte-tsdae* model slightly outperformed the hybrid model with 88.0 vs. 87.3 in PubHealthQA and 73.8 vs. 72.5 in MedQA, indicating that specific fine-tuning may be advantageous for certain tasks. Still, *med-gte-hybrid* consistently showed top performance, thus confirming that it is robust and applicable in diverse, complex tasks.

Interpretability and explainability remain core challenges when deploying advanced deep learning models in healthcare [58, 59], where sentence transformer models are no exception. Although our SHAP analysis provided insights into the most relevant features, overall model explainability remains limited. Future work should aim to improve the model's transparency, which may help to make predictions more assessable for clinicians.

Quality and content variability of clinical notes can introduce challenges in model training, as documentation practices vary between healthcare providers and institutions [60]. Moreover, differences between regional healthcare regulations, languages, and documentation standards [61] introduce challenges in medical terminology and phrasing. Clinical notes, including those found in the MIMIC-IV dataset, exhibit variability in writing habits, reflecting differences in institutional practices and EHRs [62]. While there is a shared foundation of medical terminology globally, substantial differences in phrasing, idiomatic expressions, and abbreviations persist between languages. For example, German notes are comparatively more formal and standardised than those used in the United States [61, 63]. Going forward, scalable models will need to integrate clinical texts from various healthcare providers and countries. Moreover, while sentence transformers offer superior performance in handling unstructured text, there are opportunities to further integrate structured and unstructured data sources to achieve a more holistic view of patient health. Our model could be combined with models that incorporate structured clinical data, e.g., lab results and medication codes.

# 7 Conclusion

We propose *med-gte-hybrid*, a specialised sentence transformer model tailored for the clinical domain. The *med-gte-hybrid* is designed to effectively handle long-context tasks and extract detailed narrative insights from clinical texts. *Med-gte-hybrid* consistently outperformed state-of-the-art embedding models and achieved the best overall performance in clinical prediction tasks, including CKD prognosis, eGFR prediction, and mortality prediction. Furthermore, our transformer fine-tuning approach enhanced patient stratification and facilitated the clustering of similar clinical profiles. Clustering and SHAP analyses confirmed the *med-gte-hybrid* model's ability to capture meaningful clinical narratives. Our findings highlight the potential of our clinical text fine-tuning approach to advance clinical decision support systems by integrating narrative-based insights and long-context clinical data.

# Declarations

## Ethics approval and consent to participate

MIMIC-IV datset was used for this study. For MIMIC-IV, the collection of patient information and creation of the research resource was reviewed by the Institutional Review Board at the Beth Israel Deaconess Medical Center, who granted a waiver of informed consent and approved the data sharing initiative.

## Consent for publication

Not applicable.

## Availability of data and materials

### Data availability statement

MIMIC-IV version 2.2 is available under restricted access due to privacy and ethical considerations. Access to the dataset can be requested via the PhysioNet platform [64], where applicants must complete a data use agreement and fulfil necessary requirements for access approval, including training on human subjects research. For more information, or to request access, please visit: https://physionet.org/content/mimiciv/2.2/.

### Code availability

To facilitate reproducibility and benefit other studies the source code has been shared on: https://github.com/intelligentembeddedsystemslab/med-gte-hybrid. The trained models can be made available from the authors upon reasonable request.

### Competing interests

The authors declare no competing interests.

### Funding

This work is partially funded by the German G-BA, grant number 01NVF21116.

### Authors' contributions

A.K. conceptualised the work and prepared the data. S.R. performed model fine-tuning. A.K. and S.R. drafted the initial manuscript. M.C. and O.A. provided research guidance, critically analysed results, and contributed substantially to revising the manuscript. A.K. supervised the project and administered the execution. All authors had full access to the study data, critically read and reviewed the manuscript, and approved the final version for publication.